\documentclass[runningheads]{llncs}
\usepackage[T1]{fontenc}
\usepackage{graphicx}
\usepackage{hyperref}

\usepackage{color}

\usepackage{booktabs}
\usepackage{multirow}
\usepackage{tikz}
\usepackage{amsmath}
\usepackage[binary-units]{siunitx}
\usepackage[abbreviations]{foreign}
\usepackage{xspace}
\usepackage[inline]{enumitem}
\usepackage[algo2e,ruled,lined,boxed,commentsnumbered, noend, linesnumbered, procnumbered]{algorithm2e}

\usepackage{pifont}
\newcommand{\cmark}{\ding{51}}%
\newcommand{\xmark}{\ding{55}}%

\usepackage{ifthen}
\newboolean{showcomments}
\setboolean{showcomments}{false}
\ifthenelse{\boolean{showcomments}}
{ \newcommand{\mynote}[3]{
		\fbox{\bfseries\sffamily\scriptsize#1}
		{\small$\blacktriangleright$\textsf{\emph{\color{#3}{#2}}}$\blacktriangleleft$}}
	\newcommand{\zzz}[1]{{\setlength{\fboxsep}{2pt}\fcolorbox{black}{yellow}{\textsf{\emph{#1}}}}\xspace}}
{ \newcommand{\mynote}[3]{}
	\newcommand{\zzz}[1]{}}

\usepackage{acronym}

\acrodef{DL}{decentralized learning}
\acrodef{ML}{machine learning}
\acrodef{D-PSGD}{decentralized parallel stochastic gradient descent}
\acrodef{FL}{federated learning}
\acrodef{SGD}{stochastic gradient descent}
\acrodef{IID}{independent and identically distributed}
\acrodef{non-IID}{non independent and identically distributed}
\acrodef{RMSE}{root mean square error}
\acrodef{RMW}{random model walk}
\acrodef{GL}{gossip learning}
\acrodef{EL}{epidemic learning}
\acrodef{DWT}{discrete wavelet transform}
\acrodef{FFT}{fast Fourier transform}
\acrodef{MI}{mutual information}
\acrodef{DP}{differential privacy}
\acrodef{VN}{virtual node}
\acrodef{RN}{real node}
\acrodef{LDP}{local differential privacy}
\acrodef{PNDP}{pairwise network differential privacy}
\acrodef{PNLDP}{pairwise network local differential privacy}
\acrodef{GI}{gradient inversion}
\acrodef{CML}{collaborative machine learning}
\acrodef{TPR}{true positive rate}
\acrodef{FPR}{false positive rate}
\acrodef{TEE}{trusted execution environment}
\acrodef{ANN}{approximate nearest neighbor}
\acrodef{LLM}{large language model}
\acrodef{RAG}{retrieval-augmented generation}
\acrodef{NN}{neural network}
\acrodef{MoE}{mixture of experts}
\acrodef{GNN}{graph neural network}

\newcommand{\sys}{\textsc{RAGRoute}\xspace}

\newcommand{\mmlu}{\textsc{MMLU}\xspace}
\newcommand{\medrag}{\textsc{MedRAG}\xspace}

\newcommand{\mirage}{\textsc{MIRAGE}\xspace}
\newcommand{\febrag}{\textsc{FeB4RAG}\xspace}

\newcommand{\pubmed}{\textsc{PubMed}\xspace}
\newcommand{\statpearls}{\textsc{StatPearls}\xspace}
\newcommand{\wikipedia}{\textsc{Wikipedia}\xspace}
\newcommand{\textbooks}{\textsc{Textbooks}\xspace}

\newcommand{\none}{\textsc{none}\xspace}
\newcommand{\all}{\textsc{all}\xspace}
\newcommand{\random}{\textsc{random}\xspace}
\graphicspath{ {figures/} }
\usepackage{tikz}
\usepackage[eulergreek]{sansmath}
\usepackage{pgfplots}
\usepackage{pgfplotstable}
\usepackage{cleveref}
\usepackage{comment}
\crefname{assumption}{assumption}{assumptions}

\pgfplotsset{compat=newest}
\usepgfplotslibrary{units,colorbrewer,groupplots,fillbetween,statistics}
\usetikzlibrary{patterns,shapes.misc}

\pgfplotscreateplotcyclelist{bwcool9}{
{draw=black, fill={rgb,255:red,0;green,58;blue,125},   postaction={pattern=north east lines, pattern color=black}}, 
{draw=black, fill={rgb,255:red,255;green,115;blue,182},  postaction={pattern=north west lines, pattern color=black}}, 
{draw=black, fill={rgb,255:red,0;green,141;blue,255},postaction={pattern=horizontal lines, pattern color=black}}, 
{draw=black, fill={rgb,255:red,199;green,1;blue,255},  postaction={pattern=vertical lines, pattern color=black}}, 
{draw=black, fill={rgb,255:red,249;green,232;blue,88}, postaction={pattern=dots, pattern color=black}}, 
{draw=black, fill=gray!30,                             postaction={pattern=grid, pattern color=black}}, 
{draw=black, fill={rgb,255:red,150;green,90;blue,40},  postaction={pattern=crosshatch, pattern color=black}}, 
{draw=black, fill={rgb,255:red,216;green,48;blue,52},  postaction={pattern=crosshatch dots, pattern color=black}}, 
{draw=black, fill={rgb,255:red,255;green,157;blue,58}} 
}

\newcommand{\inputplot}[2]{%
  \includegraphics{main-figure#2.pdf}%
}

\newcommand{\newgroupwidth}[2]
{\expandafter\xdef\csname groupwidth#1\endcsname{#2}}

\newcounter{groupwidth}
\newsavebox{\groupwidthbox}
\makeatletter
{\edef\groupnumber{#1}
	\stepcounter{groupwidth}%
	\@ifundefined{groupwidth\thegroupwidth}{\pgfmathsetlengthmacro{\mywidth}{\linewidth/\groupnumber}}%
	{\expandafter\let\expandafter\mywidth\csname groupwidth\thegroupwidth\endcsname}%
	\begin{lrbox}{\groupwidthbox}%
		\tikzset{/pgfplots/width={\mywidth}}%
		\ignorespaces}%
	{\end{lrbox}%
	\usebox\groupwidthbox
	\pgfmathsetlengthmacro{\mywidth}{\mywidth + (\linewidth - \wd\groupwidthbox)/\groupnumber}
	\immediate\write\@auxout{\string\newgroupwidth{\thegroupwidth}{\mywidth}}}
\makeatother

\usepackage{amsmath,amssymb,amsfonts}
\usepackage{physics}
\usepackage{enumitem}
\usepackage{thm-restate}
\usepackage{bbm}
\allowdisplaybreaks

\begin{document}
\title{Efficient Federated Search for Retrieval-Augmented Generation using Lightweight Routing}
\titlerunning{Efficient Federated Search for Retrieval-Augmented Generation}
%
\author{
Akash Dhasade\orcidID{0000-0003-4362-5548} \and
Rachid Guerraoui\orcidID{0000-0002-4794-8902} \and
Anne-Marie Kermarrec\orcidID{0000-0001-8187-724X} \and
Diana Petrescu\orcidID{0009-0006-2229-235X} \thanks{Corresponding author: diana.petrescu@epfl.ch} \and
Rafael Pires\orcidID{0000-0002-7826-1599} \and
Mathis Randl\orcidID{0009-0003-6844-4695} \and
Martijn de Vos\orcidID{0000-0003-4157-4847}
}

\authorrunning{Dhasade et al.}
%
\institute{EPFL, Lausanne, Switzerland}
%
\maketitle              

\begin{abstract}
Large language models (LLMs) achieve remarkable performance across domains but remain prone to hallucinations and inconsistencies. Retrieval-augmented generation (RAG) mitigates these issues by augmenting model inputs with relevant documents retrieved from external sources. In many real-world scenarios, relevant knowledge is fragmented across organizations or institutions, motivating the need for federated search mechanisms that can aggregate results from heterogeneous data sources without centralizing the data. We introduce \textsc{RAGRoute}, a lightweight routing mechanism for federated search in RAG systems that dynamically selects relevant data sources at query time using a neural classifier, avoiding indiscriminate querying. This selective routing reduces communication overhead and end-to-end latency while preserving retrieval quality, achieving up to 80.65\% reductions in communication volume and 52.50\% reductions in latency across three benchmarks, while matching the accuracy of querying all sources.

\end{abstract}

\section{Introduction}
\label{sec:intro}
\acresetall
\Acfp{LLM} have driven significant advancements across various domains such as natural language processing and healthcare~\cite{bharathi2024analysis,kaplan2020scaling,haltaufderheide2024ethics}. 
Despite their widespread adoption, one major concern is their tendency to \emph{hallucinate}, generating false responses with high confidence~\cite{ji2023survey} and limiting their applicability in critical domains~\cite{ji2023towards}.
\Ac{RAG} mitigates this issue by combining text generation with external retrieval, enhancing factual accuracy and contextual grounding~\cite{lewis2020retrieval,csakar2025maximizing}.

Existing \ac{RAG} systems typically rely on a single monolithic vector database~\cite{kukreja2024performance}. In practice, however, real-world knowledge is often distributed across multiple heterogeneous information systems and repositories~\cite{bhavnani2009information,wang2024feb4rag}. 
This setting calls for federated search, where queries are executed across multiple independent data sources and the results are merged into a unified ranking~\cite{shokouhi2011federated}.
In RAG systems operating over multiple repositories, federated search constitutes the retrieval layer: queries are dispatched to selected data sources, their results are aggregated and reranked, and the resulting context is passed to the language model for generation.
This avoids centralizing data, which might be complicated due to regulatory constraints or privacy considerations~\cite{callan2002distributed,kairouz2021advances}, and enables organizations to reuse existing infrastructure, therefore reducing operational and storage overhead.

A key challenge in federated search is resource selection~\cite{li2018lda,wang2024feb4rag}, i.e., identifying which sources should be queried.
Yet many \ac{RAG} pipelines query all available resources~\cite{wang2024feb4rag}.
Indiscriminate querying increases communication and computation costs~\cite{garba2020embedding} and may introduce irrelevant context that exacerbates hallucinations~\cite{bian2024influence,cuconasu2024power}.

We introduce \sys, a novel and efficient routing mechanism for federated search in RAG systems that dynamically selects relevant data sources at query time using a lightweight neural network. 
By avoiding unnecessary queries, \sys significantly reduces resource consumption and end-to-end latency while maintaining high search quality.
We evaluate \sys on three benchmarks: \mirage~\cite{xiong-etal-2024-benchmarking}, \mmlu~\cite{hendrycks2020measuring} and \febrag~\cite{wang2024feb4rag}. 
Our results show that \sys achieves up to 89.70\% recall in source selection, reduces retrieval communication volume by up to 80.65\%, and lowers end-to-end latency by up to 52.50\%, while matching the accuracy of querying all data sources.
This improvement stems primarily from reducing the number of documents that must be reranked during retrieval, thereby alleviating a major computational bottleneck in RAG pipelines.

In summary, our contributions are as follows:
\begin{enumerate}
\item We propose and implement \sys, a lightweight and effective routing mechanism for federated search in \ac{RAG} that dynamically selects data sources per query, and make our code publicly available.\footnote{See \url{https://github.com/sacs-epfl/ragroute}.}
\item We conduct extensive evaluations on three benchmarks, demonstrating that \sys significantly reduces communication overhead and latency while maintaining high retrieval quality and end-to-end accuracy.
\end{enumerate}

\section{Background and problem description}
\label{sec:prelims}

\subsection{\Acf{RAG}}
\ac{RAG} enhances the reliability of \ac{LLM} responses by integrating external information as part of the input (or \emph{prompt})~\cite{lewis2020retrieval}.
In a typical RAG pipeline, documents are split into chunks and encoded into dense vector embeddings, which are stored in a vector database that supports similarity search.
For simplicity, we refer to document chunks as documents throughout this paper.
Given a user query, the query is embedded using a compatible query encoder and a nearest-neighbor search is performed to retrieve the most relevant chunks. 
This search is often accelerated using \ac{ANN} indexing~\cite{li2019approximate}. 
The retrieved candidates are commonly reranked and appended to the original query to form an augmented prompt for the \ac{LLM}. 
By grounding generation in retrieved evidence, \ac{RAG} reduces hallucinations and improves factual accuracy without requiring model retraining.
Most existing RAG systems assume a single centralized vector database.
In contrast, we consider settings where knowledge is distributed across multiple independent data sources, motivating federated search and resource selection.

\begin{figure}[t]
	\centering
	\inputplot{plots/routing}{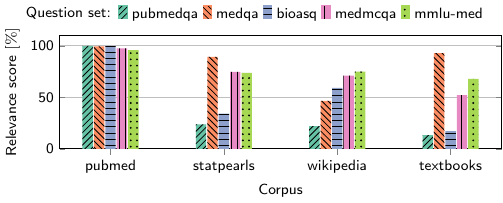}
	\caption{The relevance of different corpora in \ac{RAG} when answering questions, using question sets from the  \mirage benchmark.}
	\label{fig:routing}
\end{figure}

\subsection{Towards federated search in \ac{RAG}}
\label{subsec:motivation_towards_fedrag}
Federated search is an information retrieval setting in which a query is executed across multiple independent data sources and the results are aggregated without centralizing the underlying data~\cite{shokouhi2011federated}.
A key challenge in federated search is resource selection, i.e., determining which data sources should be queried~\cite{wang2024feb4rag}.
In \ac{RAG} systems operating over multiple repositories, source relevance varies substantially across queries, making accurate relevance estimation essential for efficient retrieval.

We empirically show this by analyzing data source relevance using corpora and questions from the \mirage benchmark (more details can be found in \Cref{sec:experimental_setup}).
This benchmark contains a large number of medical multiple-choice questions and answers and is divided into five question sets~\cite{xiong-etal-2024-benchmarking}.
As knowledge backend for \ac{RAG} we use four different data sources (corpora), namely \pubmed, \statpearls, \wikipedia and \textbooks.
For each question, we determine which corpora are relevant by considering a corpus relevant if at least one document originating from that corpus appears in the top-15 retrieved results.

\Cref{fig:routing} shows the overall relevance of different corpora, highlighting how corpus usefulness varies across question sets.
For example, the bar corresponding to the \textsc{MedQA} question set and the \textsc{StatPearls} corpus shows a relevance score of 89.32\%, meaning that for 89.32\% of queries in \textsc{MedQA}, at least one document in the retrieved relevant documents originates from \textsc{StatPearls}.
While some corpora, such as \textsc{PubMed}, consistently provide valuable, relevant information for all question sets, relying on a single corpus is often insufficient.
Indeed, results from~\cite{xiong-etal-2024-benchmarking} demonstrate that combining multiple corpora improves retrieval performance.
Some corpora, such as \statpearls or \wikipedia, are only useful in particular cases.
The \textbooks corpus, for example, is mostly irrelevant for the \textsc{PubMedQA} question set.
The differences in corpus relevance motivate the importance of adequate resource selection for a given query.

One must strike a balance in the number of data sources being queried.
While querying all available data sources guarantees full coverage, it also increases retrieval latency and computational overhead, as more requests, database searches, and document reranking operations are required.
At the same time, under-selecting data sources risks missing critical information, particularly in domains where information is distributed sparsely across multiple repositories, for example, government data that resides in different portals~\cite{purdueFederatedSearch2010}.
Achieving a good trade-off between retrieval efficiency and response quality remains an open problem.
Therefore, this work answers the following question: \emph{how can we efficiently predict query-specific source relevance in federated search for \ac{RAG}, while minimizing retrieval overhead?}

\begin{figure*}[t!]
    \centering
    \includegraphics[width=.95\linewidth]{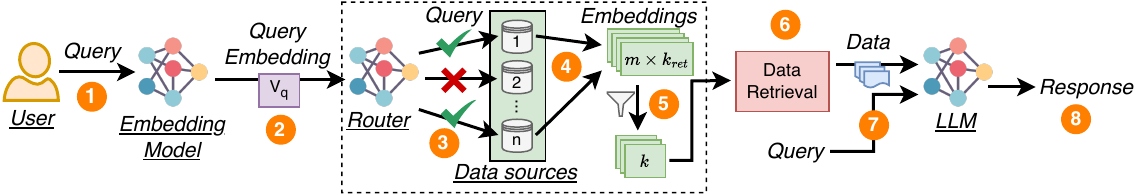}
    \caption{The workflow of \sys. The components specific to \sys are indicated in the box with the dashed border.
    }
    \label{fig:workflow}
\end{figure*}

\section{Design of \sys}
\label{sec:design}

\subsection{System model and assumptions}
\label{sec:assumptions}
We assume a permissioned setting in which all data sources are known and trusted.
Thus, we focus on federated search within an enterprise or institutional consortium.
Each data source maintains its own data and associated local embeddings, and is responsible for computing and storing vector representations of its documents.
The specifications of the embedding models used could vary across data sources.
We assume that the enterprise or institution running the system has access to the embedding models used by the different data sources.
User queries are submitted in natural language and converted into embedding vectors using the appropriate model(s).
While we assume data sources are generally available, \sys remains functional even if some sources are temporarily offline.
The system can simply exclude the offline data source from selection, ensuring graceful degradation in retrieval coverage rather than system failure.

\subsection{\sys workflow}
\label{sec:workflow}
We visualize the \sys workflow in~\Cref{fig:workflow}, enabling \ac{RAG}-enhanced \ac{LLM} responses by retrieving documents from $ n $ distinct data sources.
We show the components specific to \sys in the dashed box.
When a user sends a query to the system~{\color{orange}\ding{202}}, the user query is first converted into one or more embeddings using all embedding models used by the data sources~{\color{orange}\ding{203}}.
These query embeddings are then forwarded to a \emph{router}, whose purpose is to decide which of the $ n $ data sources are relevant.
The router predicts this relevance separately for each source using source specific features and query embeddings.
We outline the design and operation of our router in \Cref{s:query-routing}.

After determining the $ m $ relevant data sources ($ m \leq n $), we forward to each selected source the query embedding previously computed using its respective embedding model~{\color{orange}\ding{204}}.
For example, \Cref{fig:workflow} shows that data source 1 is selected and receives the query embedding, while source 2 is skipped.
Each selected source uses the received query embedding to retrieve top-$k_{ret}$ documents most similar to the query.
This results in $ m \times k_{ret} $ total retrieved documents~{\color{orange}\ding{205}}.
This is followed by a post-retrieval reranking step, which has become a standard component in modern RAG systems~\cite{gao2024retrievalaugmentedgenerationlargelanguage,pinecone2024rerank,openai2024rerank}.
Rerankers, often implemented as cross-encoder models, rescore the retrieved candidates to prioritize the passages most semantically relevant to the user query.
This two-stage retrieval design improves grounding accuracy while also reducing response latency, since only a compact set of highly relevant documents is passed to the language model.
After reranking, only final $k$ documents are retained where the value of $k$ can differ from the retrieval cutoff $k_{ret}$ (see \Cref{sec:evaluation}).

Finally, the relevant documents~{\color{orange}\ding{207}} and original user query~{\color{orange}\ding{208}} are combined into a single prompt.
This prompt is fed to the \ac{LLM} and a response is generated and returned to the user~{\color{orange}\ding{209}}, therefore completing the query.

\subsection{Lightweight query routing}
\label{s:query-routing}
To enable resource selection and efficient retrieval across multiple data sources, \sys uses a lightweight query router, implemented as a shallow \ac{NN} with a few fully connected layers.
This minimal design is intentional: the router must remain computationally inexpensive so that routing overhead is negligible compared to retrieval and generation.
Despite its simplicity, the router is sufficient to estimate the relevance of each data source before retrieval.
Using a shallow \ac{NN} is inspired by practices in \ac{MoE} models and ensembles.
\ac{MoE} models leverage a small router function to decide which subset of experts to activate~\cite{zhou2022mixture}.
Similarly, shallow \acp{NN} are used for decision-making in one-shot federated ensembles~\cite{allouah2025revisiting}.
This work applies similar ideas to selecting relevant data sources in federated search for \ac{RAG} systems.
We next describe the training and inference phase of the \sys router.

\subsubsection{Training phase}
\label{subsubsec:training_phase}
Let \( R = \{R_1, R_2, \ldots, R_n\} \) denote the set of \( n \) data sources, where each \( R_i \) corresponds to a collection of documents. 
For each query–source pair \( (q, R_i) \), the router estimates a relevance probability, which is thresholded to obtain a binary routing decision.
The router is trained using ground-truth binary relevance labels 
\( s(q, R_i) \in \{0,1\} \), 
where \( s(q, R_i) = 1 \) indicates that source \( R_i \) is relevant to query \( q \).
We first describe how these ground-truth labels are constructed, before detailing the router input features. 
We consider two approaches for constructing the source-level relevance labels.

\begin{enumerate}
    \item \textbf{Rerank based} -- For a given query \( q \), the top-\( k_{\text{ret}} \) documents are retrieved from each source \( R_i \in R \).
    All retrieved documents are then reranked jointly using a neural reranker to produce a global top-\( k \) list.
    A source \( R_i \) is labeled as relevant if at least one retrieved document \( d \) from \( R_i \) appears in the global top-\( k \):    \[
    s(q, R_i) = 
    \begin{cases}
        1 & \text{if } \exists \text{ } d \in R_i \text{ s.t. } d \text{ is in the global top-}k \\
        0 & \text{otherwise.}
    \end{cases}
    \]
    These labels depend on both the embedding model and the value of $k$.
    During inference, we use the same value of $k$ as for training.

    \item \textbf{\ac{LLM} based}~\cite{wang2024feb4rag} -- For each query \( q \), we first retrieve the the top-\( k_{\text{ret}} \) documents from each source \( R_i \in R \).
    We then obtain query–document relevance judgments for these retrieved documents using an external \ac{LLM} that is independent of the embedding model.
    Each document is assigned one of four labels: not relevant, minimally relevant, highly relevant, or key, where key indicates a strong match.
    For each source \( R_i \), we aggregate the LLM judgments of its retrieved documents into a graded precision score:
    \[
        \text{Graded Precision}(q, R_i) = \frac{\sum_{j=1}^{k} w(q, d_j)}{k} \times 100
    \]
    where \( d_j \) denotes the \( j \)-th retrieved document from source \( R_i \) for query \( q \), and \( w(q,d_j) \) is defined as:

    \[
    w(q, d_j) =
    \begin{cases} 
        0       & \text{if not relevant} \\
        0.25    & \text{if minimally relevant} \\
        0.5     & \text{if highly relevant} \\
        1       & \text{if key.}
    \end{cases}
    \]

    Finally, a source is labeled as relevant if the Graded Precision score is positive:

    \[
    s(q, R_i) = 
    \begin{cases}
        1 & \text{Graded Precision}(q, R_i) > 0 \\
        0 & \text{otherwise.}
    \end{cases}
    \]
    These labels depend only on the cutoff \( k_{\text{ret}} \) and not on the embedding model.
\end{enumerate}

\textbf{Feature selection.} While it is common to assume that each source uses the same embedding model to embed its documents~\cite{gnn_learn_to_rank}, some specialized sources may instead employ their own embedding model~\cite{wang2024feb4rag}.
We design our router to support the more general scenario where each source could have its own embedding model. 
Let $H_i(x) \in \mathbb{R}^{z_i}$ denote the embedding of any input $x$ (a query or a document) using the embedding model of source $R_i$, where $z_i$ is its embedding dimension. 
The router takes the following three features as input:
\begin{enumerate}[label=\emph{(\roman*)}]
\item the query embedding $H_i(q)$, 
\item the centroid of the data source $C_i = \frac{1}{|R_i|}\sum_{d \in R_i}H_i(d)$, and
\item the source-id as one hot encoded vector $\text{Id}_i$.
\end{enumerate}
The centroid $C_i$, computed as the average vector representation of all document embeddings in a data source, summarizes its overall semantic content.
The source-id serves as a prior signal to help the router account for systematic differences across sources.
Since the size of the embedding $z_i$ may differ across sources, we consider the highest embedding size $z = \max_{i \in [n]} z_i$ and pad zeros if $z_i < z$.
We denote the padded query embedding as $\hat{H}_i(q)$ and the padded centroid as $\hat{C}_i$, where $\hat{H}_i(q), \hat{C}_i \in \mathbb{R}^z$.
The router parameterized by $\theta$ and denoted by $f_\theta$ independently predicts a relevance probability for each source $i \in [n]$ based on these features.
Given a dataset of queries $\mathcal{D}_\textrm{train}$ with ground truth relevance labels constructed as discussed before, the router is trained to minimize the following objective:
\begin{equation}
\mathcal{L}(\theta) = \sum_{q \in \mathcal{D}_\textrm{train}} \sum_{i = 1}^n 
\ell\left( f_\theta\bigl( \hat{H}_i(q), \hat{C}_i, \text{Id}_i \bigr),\ s\bigl(q, R_i\bigr) \right)
\end{equation}
where $\ell$ is a binary classification loss, such as binary cross-entropy.

\subsubsection{Inference phase}
Once trained, \sys uses this model to efficiently route incoming user queries to relevant data sources.
We run one forward pass for each of the available data sources individually to predict their relevance to a given inference query.
This forward pass completes quickly (with sub-millisecond latency, see \Cref{sec:exp_efficiency}) and can be done in parallel for individual sources. 
Additionally, multiple queries can be batched into a single forward pass, depending on their arrival time.
When new data sources are added or existing ones are updated, \sys regenerates the training ground truth by querying the affected sources together with those predicted as relevant by the existing router.
This targeted querying strategy ensures that new and updated sources are incorporated into the label construction process while avoiding unnecessary queries to unrelated ones, thereby minimizing update overhead.
Because the router is implemented as a shallow \ac{NN} with only a few fully connected layers, retraining is highly lightweight, requiring minimal computation and storage. 
Thus, the router can be rapidly retrained in the background whenever updates occur.

\section{Evaluation}
\label{sec:evaluation}

\subsection{Experimental setup}
\label{sec:experimental_setup}

\textbf{Implementation.}
We implement \sys in Python using an event-driven architecture based on \texttt{asyncio}.
Each core component (coordinator, router, data sources, and LLM engine) runs as an independent process to enable modularity and parallel execution.
The coordinator orchestrates asynchronous communication across components.
We use the \textsc{ZeroMQ} library for inter-process messaging and \textsc{AIOHTTP} to handle incoming HTTP queries.
We use the \textsc{Ollama} framework for inference which provides a convenient way to load and run inference with different \acp{LLM}~\cite{ollama}.
For the embedding models, we use the \textsc{PyTorch} library.

\textbf{Router model.}
We implement the router as a lightweight fully connected \ac{NN}.
The network consists of hidden layers with 128, 64, and 32 neurons, each followed by Layer Normalization, ReLU activation, and Dropout to improve stability and prevent overfitting.
These hyperparameters were selected through cross-validation, where we evaluated several architectures with varying numbers of layers and hidden dimensions on the validation set.
The output layer consists of a single neuron that produces a raw logit score, predicting whether the corpus is relevant to the given query.
The model is trained using Binary Cross-Entropy with Logits Loss with a positional weight to address class imbalance.
We use a cyclic scheduler for the learning rate $ \gamma $, oscillating $\gamma$ between 0.001 and 0.005.
Model performance is evaluated on the validation set after each epoch, and the best model is selected based on validation accuracy.
Training data are split by question into 30\%/10\%/60\% train/validation/test partitions, and all input features are standardized using a \textsc{StandardScaler}.
The router’s small size ensures fast training, negligible inference overhead, and ease of retraining when data sources evolve.
We also tested alternative classifiers (\eg, logistic regression and random forests) but found the shallow \ac{NN} performed best overall.

\textbf{Datasets.}
We evaluate \sys with the following three benchmarks:

\begin{enumerate}[label=\emph{(\roman*)}]

\item \textbf{\mirage} is a benchmark designed to evaluate \ac{RAG} systems for medical question answering~\cite{xiong-etal-2024-benchmarking}.
It consists of \num{7663} questions drawn from five widely used medical QA datasets.
We use \medrag as knowledge source, which includes four corpora with documents related to healthcare~\cite{xiong-etal-2024-benchmarking}.
For generating embeddings, we use \textsc{MedCPT}~\cite{jin2023medcpt}, a domain-specific model designed for biomedical contexts.
For retrieval, we use the \textsc{IndexFlatL2} index structure, provided by the \textsc{FAISS} library~\cite{douze2024faiss}, ensuring exact search and eliminating sources of approximation in our experiments.
We treat each corpus as a separate data source.
For \mirage, we construct the ground truth relevance labels using the rerank based approach.
To run \sys with a \ac{RAG} pipeline, we leverage the code provided by the \medrag toolkit.

\item \textbf{\mmlu} is a benchmark that evaluates \ac{LLM} systems across tasks ranging from elementary mathematics to legal reasoning~\cite{hendrycks2020measuring}.
For our experiments, we use eight subject-specific subsets of \mmlu with a total of \num{2803} questions.
As a knowledge source, we use a Wikipedia dataset~\cite{karpukhin-etal-2020-dense}.
From this dataset, we cluster the documents into ten groups using the $k$-means algorithm to simulate different data sources.
After clustering, we observe variance in the cluster size, ranging from 1.41~M to 2.88~M vectors per cluster.
For \mmlu, we construct the ground truth relevance labels using the rerank based approach.
To run \mmlu, we leverage the code provided by the \textsc{RQABench} framework~\cite{retrieval_qa_benchmark}.

\item \textbf{\febrag} is a benchmark designed to evaluate federated search methods for \ac{RAG} systems~\cite{wang2024feb4rag}. 
It consists of \num{790} user queries spanning diverse domains and complexity levels. 
\febrag is derived from \textsc{BEIR}~\cite{thakur2beir} and includes 13 heterogeneous data sources powered by eight distinct embedding models, enabling evaluation under realistic federated retrieval settings.
For \febrag, ground truth relevance labels are obtained using the \ac{LLM} based approach.
Unlike \mirage and \mmlu, \febrag does not provide verifiable ground-truth answers.
\end{enumerate}

\textbf{Evaluation.}
To facilitate automated evaluation, we developed a separate benchmarking script that iterates over all questions in a given dataset and sends each query, along with the associated answer choices if applicable, to the \sys system via HTTP requests.
Queries are sent one by one: the script waits for the response to a given query before proceeding to the next.
Upon receiving a response from the system, the script verifies the correctness of the answer against the ground truth answer, if applicable.
This setup enables systematic and reproducible evaluation across multiple benchmarks.

\textbf{Retrieval and reranking.}
For all datasets, we retrieve a global top-$k=15$ list of documents for generation.
To construct this final set, each selected data source retrieves top-$k_{ret}=50$ documents most similar to the query using exact similarity search with \textsc{FAISS}~\cite{douze2024faiss} incurring negligible latency compared to the reranker.
We set $k_{ret}$ sufficiently large to ensure high recall, while balancing the trade-off with reranking cost.
All retrieved candidates are then reranked to produce the global top-15 list.
This two-stage retrieval strategy balances recall and precision: a sufficiently large retrieval pool ensures coverage, while reranking improves semantic relevance and reduces noise in the final context.
We employ the \texttt{BAAI/bge-reranker-v2-m3} model~\cite{BAAI_bge_reranker_v2_m3}, a lightweight cross-encoder reranker designed for multilingual reranking with efficient inference.

\textbf{\ac{LLM} models.}
As \ac{LLM}, we use the open-source LLaMA 3.1 8B Instruct model for all above datasets~\cite{dubey2024llama} as it is commonly considered in related work~\cite{addison2024c,salve2024collaborative}. We adopt a zero-shot chain-of-thought prompting scheme, instructing the model to reason step-by-step before providing the final answer. The output is formatted in JSON to ensure interpretable reasoning and structured evaluation.

\textbf{Hardware.}
We run our experiments on a compute cluster equipped with an NVIDIA A100 GPU for \ac{LLM} answer generation, and 500 GB of main memory.

\begin{table*}[t]
\centering
\caption{Classification metrics (averages) for our router and for different benchmarks. \sys router achieves high accuracy and recall, demonstrating good generalization across benchmarks.}
\label{tab:summary_classification_results}
\begin{tabular}{lcccccc}
\toprule
\textbf{Benchmark} & \textbf{Accuracy (\%)} & \textbf{Precision (\%)} & \textbf{Recall (\%)} & \textbf{F1-Score (\%)} & \textbf{AUC (\%)} \\
\midrule
\mirage & 86.63 & 86.79 &  83.35 & 84.96 & 92.94 \\
\mmlu & 90.93 & 71.64 & 82.92 & 76.87 & 95.77 \\
\febrag & 83.05 & 87.37 & 89.70 & 88.51 & 84.00 \\
\bottomrule
\end{tabular}
\end{table*}

\textbf{Routing baselines.}
To analyze the effectiveness and efficiency of \sys, we experiment with the following four routing strategies.
\begin{enumerate}[label=\emph{(\roman*)}]
    \item \textbf{\none.} This routing strategy does not query any data source and the input prompt to the \ac{LLM} is not enhanced with retrieved documents.
    \item \textbf{\all.} Under this routing strategy, all data sources are queried. This can be considered as a naive baseline for federated search that lacks a mechanism for strategic resource selection.
    \item \textbf{\sys.} This routing strategy uses the \sys router to identify and retrieve documents only from relevant data sources.
    \item \textbf{\random.} This routing strategy randomly selects a fixed number of data sources, matching the number selected by \sys but without using relevance predictions.
    We incorporate this strategy to evaluate the effectiveness of \sys beyond extremes like querying all or no data sources.
    This allows us to ignore the effect of the number of contacted sources, isolating the impact of how the sources are selected, demonstrating that \sys gains arise from intelligent, query-aware routing.
\end{enumerate}

\textbf{Metrics.}
Our experiments primarily focus on the classification performance of the \sys router model and the system efficiency of the entire \sys system.
For the former, we report standard classification metrics such as accuracy, recall, precision, F1-Score and AUC.
For the latter, we monitor, for each user query, relevant system metrics such as the number of data sources contacted, communication volume and latency.
We also determine the end-to-end \ac{RAG} accuracy for the \mmlu and \mirage benchmarks.
We are unable to do so for \febrag since this benchmark does not provide ground-truth answers.

\subsection{\sys routing effectiveness}
\label{sec:exp_routing_effectiveness}
We evaluate the effectiveness of our router and show its classification performance in predicting data source relevance for a given query in the test set for each benchmark. 
\Cref{tab:summary_classification_results} presents various classification metrics, \ie, accuracy, precision, recall, F1-score, and AUC, for all three benchmarks.
Here, recall measures the router’s ability to identify all relevant data sources, while accuracy reflects the overall correctness of the router’s binary predictions (and not the end-to-end \ac{LLM} accuracy in generating final responses).

We achieve consistently strong results across all benchmarks, with accuracy ranging from 83.05\% for \febrag to 90.93\% for \mmlu, and recall values between 89.70\% and 82.92\% respectively, indicating that the router reliably identifies relevant data sources across diverse settings.
The slightly lower accuracy on some benchmarks primarily stems from our design choice to favor recall which is particularly important for imbalanced datasets where only a few sources are relevant per query.
This trade-off is desirable in federated retrieval settings, where missing a relevant data source is typically more detrimental to the quality of \ac{LLM} answers than querying an additional one.
Overall, the \sys router demonstrates strong and balanced generalization across benchmarks, confirming its effectiveness for real-world federated search in \ac{RAG} systems.

\begin{figure}[t]
	\centering
	\inputplot{plots/communication}{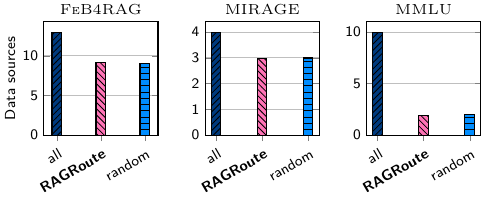}
	\caption{The average number of queries (data sources contacted) for all benchmarks and for different routing strategies. \sys significantly reduces the number of contacted data sources compared to \textsc{all} baseline.}
	\label{fig:communication}
\end{figure}

\begin{figure}[t]
    \centering
    \resizebox{0.98\linewidth}{!}{%
    \inputplot{plots/chart-matrix}{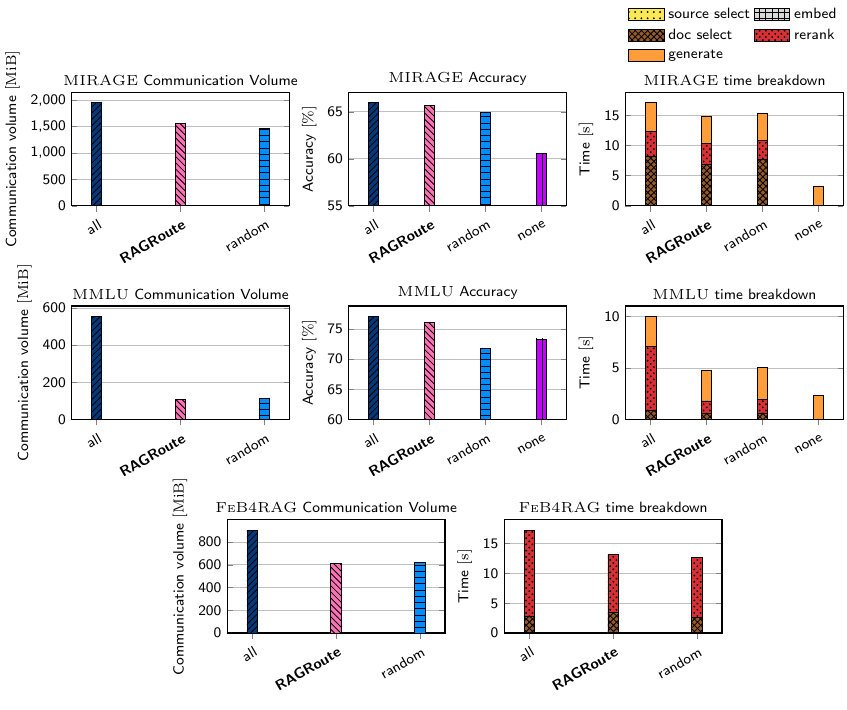}
    }
    \caption{The communication volume required for document retrieval (left), \ac{RAG} test accuracy (middle) and query time breakdown (right), for all routing strategies and benchmarks. \sys consistently reduces communication volume and maintains near-optimal accuracy, with improved query latency.
    }
    \label{fig:chart_matrix}
\end{figure}

\subsection{\sys efficiency gains}
\label{sec:exp_efficiency}
We now quantify the reduction by \sys in the number of data sources contacted and communication volume related to document retrieval for all routing baselines.
We also analyze the end-to-end \ac{RAG} accuracy and provide a time breakdown of different operations in the \ac{RAG} workflow.

\subsubsection{Number of data sources contacted}
\label{sec:exp_data_sources_contacted}
\Cref{fig:communication} shows the total number of data sources contacted across all queries, for all routing strategies and benchmarks.
We find that the number of data sources contacted for the \sys routing strategy is always lower compared to querying all data sources (which is the \textsc{all} routing strategy).
This effect is the most pronounced on the \mmlu dataset, where the number of contacted data sources decreases from \num{16810} to \num{3250}, representing an \emph{80.67\% reduction} in the number of messages exchanged for document retrieval.
In other words, under the \sys routing strategy, only 1.93 out of ten data sources are contacted on average.
On average, \sys contacts 2.98 out of four data sources per query on \mirage, and 9.20 out of thirteen on \febrag.
These results highlight the effectiveness of \sys in minimizing communication and computation overhead during federated retrieval, while maintaining high routing accuracy.

\subsubsection{Communication volume}
We next show the reduction in communication volume achieved by querying only relevant data sources.
By selecting a subset of sources predicted as relevant, \sys significantly decreases the communication volume between the coordinator and data sources.
\Cref{fig:chart_matrix} (left column) reports the total communication volume for the three routing strategies: \all, \sys, and \random.
Compared to the \all baseline, \sys reduces total communication volume by 19.94\% (\SI{1955.4}{\mebi\byte} $\rightarrow$ \SI{1565.4}{\mebi\byte}) on \mirage, by 80.65\% (\SI{554.4}{\mebi\byte} $\rightarrow$ \SI{107.3}{\mebi\byte}) on \mmlu, and by 32.52\% (\SI{905.3}{\mebi\byte} $\rightarrow$ \SI{610.9}{\mebi\byte}) on \febrag.
The \random baseline, which queries a similar number of sources as \sys, also lowers communication volume compared to \all, but at the cost of lower accuracy, as shown next.

\subsubsection{End-to-end \ac{RAG} accuracy}
Finally, we report the average end-to-end accuracy on the \mirage and \mmlu benchmarks.
For the \febrag benchmark, this metric is omitted, as no ground-truth answers are available to evaluate the correctness of generated outputs.

\Cref{fig:chart_matrix} (middle column) shows the end-to-end accuracy results.
On the \mmlu benchmark, \sys achieves an average accuracy of 76.09\%, nearly matching the \all baseline (77.10\%).
In contrast, the \random baseline reaches only 71.74\%, underperforming even the \none baseline (73.35\%).
This drop illustrates the negative impact of indiscriminate retrieval: introducing irrelevant or noisy documents can distract the language model and ultimately reduce answer quality.
By contrast, \sys’s selective routing ensures that documents only from relevant sources contribute to the answer quality.

On the \mirage benchmark, the \none baseline achieves 60.58\% accuracy, while randomly querying data sources increases it to 64.98\%.
When the data sources are selected via \sys, accuracy further improves to 65.64\%, closely matching the \all baseline at 66.96\%.
Overall, \sys maintains nearly optimal \ac{RAG} accuracy while greatly reducing communication volume.

\subsubsection{Time breakdown}
We further explore the efficiency gains of \sys and provide for each dataset a time breakdown when answering a user query.
These results are shown in \Cref{fig:chart_matrix} (right column) for each of the routing baselines, and we measure the time spent in the following five components.
\textsc{Source Selection} refers to the time spent on the inference request of the \sys router, which predicts the set of relevant data sources for a given query.
\textsc{Embedding} denotes the time required to compute the query embeddings using the appropriate embedding models for all data sources.
\textsc{Doc Selection} measures the time elapsed from when the coordinator dispatches the query to the selected data sources until all retrieval results are received.
This includes network communication, client-side retrieval, and result transmission.
\textsc{Rerank} represents the time consumed by the reranker to rescore and reorder the retrieved candidates, producing the global top-$k$ set of documents.
Finally, \textsc{Generate} corresponds to the time required by the \ac{LLM} to synthesize the final response given the reranked context documents and user query.
The \textsc{Source Selection}, \textsc{Embedding}, \textsc{Doc Selection}, and \textsc{Rerank} times are zero for the \textsc{None} routing baseline, as no data sources are queried in this configuration.
For the \febrag benchmark, we report only the \textsc{Source Selection}, \textsc{Embedding}, \textsc{Doc Selection}, and \textsc{Rerank} times, since no ground-truth answers are available to evaluate the \textsc{Generate} phase.

We first observe that the end-to-end query latency of \sys varies across datasets, reflecting differences in corpus scale.
When using the \sys routing strategy, queries complete on average in \SI{14.91}{\second} on \mirage and \SI{4.75}{\second} on \mmlu.
These variations are explained by two factors.
First, \mirage contains larger and more textually rich corpora, increasing the time needed to fetch and process documents.
Second, the number of tokens included in the \ac{LLM} input prompt is approximately twice as high for \mirage compared to \mmlu, resulting in longer generation times (\SI{4.63}{\second} vs. \SI{2.96}{\second}).
For \febrag, the query answer latency averages around \SI{13.21}{\second}, but this value excludes generation time because no verifiable ground-truth answers are available for this dataset.

An interesting observation is that reranking is a major contributor to overall latency.
For example, in the \textsc{All} baseline, reranking alone accounts for \SI{6.27}{\second} on \mmlu (around 62.64\% of total latency) and \SI{14.49}{\second} on \febrag (around 83.90\% of the retrieval pipeline, excluding generation), showing that the cross-encoder reranker can become a major computational bottleneck in \ac{RAG} systems operating over multiple data sources, as the number of considered sources grows.
In contrast, \sys effectively mitigates this bottleneck by fetching documents from fewer sources, effectively reducing the number of candidate documents that must be reranked.
With \sys, reranking time decreases by 80.70\% on \mmlu (from \SI{6.27}{\second} to \SI{1.21}{\second}) and by 32.51\% on \febrag (from \SI{14.49}{\second} to \SI{9.78}{\second}).
Importantly, this reranking cost reduction directly translates into faster end-to-end query execution, without compromising retrieval accuracy or grounding quality.

Meanwhile, the latency overhead of pre-retrieval components (embedding generation and routing inference) is almost negligible.
Even for \febrag, where the query must be embedded multiple times using different models, the overall embedding and routing time remains imperceptible on the figure.
In standalone inference measurements with a batch size of 32, the router imposes an average latency of only \SI{0.4}{\milli\second} using an NVIDIA A100 GPU and \SI{0.8}{\milli\second} using an AMD EPYC 7543 32-Core CPU.
This highlights that the routing step performed by \sys adds negligible computational overhead and has an insignificant impact on the end-to-end query latency.
Overall, these results show that the pre-retrieval routing in \sys significantly alleviates the reranking bottleneck while adding almost no overhead.
Our design enables scalable, low-latency federated retrieval even when the number of data sources grows.

\subsection{Ablation study}
\label{sec:exp_ablation}
We conduct an ablation study to evaluate the contribution of different input features to the router's performance on the \mirage and \febrag benchmarks. Specifically, we train the router to predict relevance using the following combinations of features: \emph{(i)} the query embedding and the centroid, \emph{(ii)} the query embedding and the source-id, and \emph{(iii)} all three features (\Cref{subsubsec:training_phase}).
The query embedding must always be present as an input feature to be able to predict relevance for that query.
\Cref{tab:ablation} summarizes the results.
The model trained with all three features achieves the highest recall on both the \mirage and \febrag benchmarks, indicating that both the centroid and the source-id are important for effective routing.
We also explored additional features such as the number of documents per source and the density around the centroid.
However, including these features did not lead to further performance improvements.

\begin{table}[t!]
    \centering
    \small
    \caption{Ablation study of router's input features. Using all three features results in the highest performance on both benchmarks.}
    \label{tab:ablation}
    \begin{tabular}{c c c c c}
    \toprule
       \multirow{2}{*}{query}  & \multirow{2}{*}{centroid} & \multirow{2}{*}{source-id} & \multicolumn{2}{c}{Recall} \\
       \cmidrule(lr){4-5}
       & & & \mirage & \febrag \\
       \midrule
       \cmark  &  \cmark & \xmark & $81.53$ & $89.07$\\
       \cmark  &  \xmark & \cmark & $82.33$ & $88.33$\\
       \cmark  &  \cmark & \cmark & $\textbf{83.35}$ & $\textbf{89.70}$\\
       \bottomrule
    \end{tabular}
\end{table}

\section{Related work}
\label{sec:related}

\textbf{\ac{RAG} with multiple data sources.}
\textsc{FeB4RAG} examines federated search within the \ac{RAG}  paradigm and focuses on optimizing resource selection and result merging to enhance retrieval efficiency~\cite{wang2024feb4rag}.
The underlying idea consists of introducing a dataset for federated search and incorporating \ac{LLM}-based relevance judgments to benchmark resource selection strategies.
Notably, the paper emphasizes the importance of developing novel federated search strategies for \ac{RAG}. 
Salve et al. propose a multi-agent \ac{RAG} system where different agents handle the querying of databases with differing data formats (\eg, relational or NoSQL)~\cite{salve2024collaborative}.

Other approaches focus on privacy in federated search.
\textsc{Raffle} is a framework that integrates \ac{RAG} into the federated learning pipeline and leverages public datasets during training while using private data only at inference time~\cite{muhamedcache}.
\textsc{C-FedRag} is a federated \ac{RAG} approach that enables queries across multiple data sources and leverages hardware-based \acp{TEE} to ensure data confidentiality~\cite{addison2024c}.
\textsc{FRAG} leverages homomorphic encryption to enable parties to collaboratively perform \ac{ANN} searches on encrypted query vectors and data stored in distributed vector databases, ensuring that no party can access others' data or queries~\cite{zhao2024frag}.
These schemes can benefit from \sys while ensuring privacy-preserving federated search.

\textbf{\Ac{ML}-assisted resource selection.}
\ac{ML} models have been explored to support resource selection in federated search~\cite{garba2023federated}.
Arguello \etal leverage different features, \eg, the topic of queries, and train a classifier for resource selection~\cite{arguello2009classification}.
Learn-to-rank approaches such as \textsc{SVMrank}~\cite{dai2017learning} and the LambdaMART-based \textsc{LTRRS}~\cite{wu2019ltrrs} refine relevance rankings by leveraging diverse feature sets.
Ergashev \etal construct a heterogeneous graph to capture query-source and source-source relationships and then predict the query-source relevance ranking using a \ac{GNN}~\cite{gnn_learn_to_rank}.
Wang \etal use an \ac{LLM} as a resource selector, introducing a novel prompting approach called ReSLLM~\cite{wang2024resllm}. 
They also propose to fine-tune ReSLLM through previously logged queries and snippets from data sources.
However, these approaches are either more computationally expensive than the lightweight \sys router or cannot handle heterogeneous embedding models across data sources.

\section{Conclusion}
\label{sec:conclusion}
We presented \sys, a novel and efficient routing mechanism for federated search in \ac{RAG} systems. 
By dynamically choosing relevant data sources at query time via a lightweight neural classifier, \sys minimizes unnecessary queries while preserving high retrieval quality. 
Evaluations on \mirage, \mmlu, and \febrag demonstrate that \sys reduces the number of contacted sources and document retrieval communication volume by up to 80.65\%, and decreases end-to-end latency by up to 52.50\%, with minimal impact on end-to-end accuracy. 
These gains are primarily achieved by reducing reranking overhead which constitutes a major bottleneck in RAG pipelines. 
Our results confirm that querying all data sources is often unnecessary, underscoring the importance of query-aware retrieval strategies in federated search workflows for \ac{RAG}.

%
%
%
%




\newpage
\bibliographystyle{splncs04}
\bibliography{main.bib}
\end{document}